\pdfoutput=1

\documentclass[10pt,twocolumn,letterpaper]{article}

\usepackage{iccv}
\usepackage{times}
\usepackage{epsfig}
\usepackage{graphicx}
\usepackage{amsmath}
\usepackage{amssymb}

\newcommand{\blockcomment}[1]{}

\usepackage{enumitem}

\usepackage{subcaption}
\usepackage{graphicx, trimclip}
\newcommand\picdims[4][]{%
  \setbox0=\hbox{\includegraphics[#1]{#4}}%
  \clipbox{.5\dimexpr\wd0-#2\relax{} %
           .5\dimexpr\ht0-#3\relax{} %
           .5\dimexpr\wd0-#2\relax{} %
           .5\dimexpr\ht0-#3\relax}{\includegraphics[#1]{#4}}}

\usepackage{tikz}
\usepackage{pgfplots}
\usetikzlibrary{pgfplots.statistics}
\usepgfplotslibrary{colorbrewer}
\pgfplotsset{compat = 1.15, cycle list/Set1-8}
\usepackage{adjustbox}
\usetikzlibrary{plotmarks}

\usepackage{xcolor}
\definecolor{linecolor}{RGB}{96, 96, 96}
\definecolor{colorone}{RGB}{177,186,207}
\definecolor{colortwo}{RGB}{208,212,226}
\definecolor{colorthree}{RGB}{239,239,245}
\definecolor{colorfour}{RGB}{225,175,168}
\definecolor{colorfive}{RGB}{239,207,201}
\definecolor{colorsix}{RGB}{247,235,233}

\definecolor{morecontrast-one}{RGB}{98,103,115} 
\definecolor{morecontrast-two}{RGB}{138,151,177}
\definecolor{morecontrast-three}{RGB}{224,227,236}
\definecolor{morecontrast-four}{RGB}{95,63,63}
\definecolor{morecontrast-five}{RGB}{193,111,102}
\definecolor{morecontrast-six}{RGB}{233,191,185}

\usepackage{tabularx}
\usepackage{booktabs}
\usepackage{array}
\newcolumntype{x}[1]{%
>{\raggedleft\hspace{0pt}}p{#1}}%
\newcommand{\tn}{\tabularnewline}
\usepackage{bm}

\usepackage{stfloats}

\usepackage[pagebackref=true,breaklinks=true,bookmarks=false, hidelinks]{hyperref}
\renewcommand*{\backrefalt}[4]{%
    \ifcase #1 (Not cited.)%
    \or        (Cited on page~#2)%
    \else      (Cited on pages~#2)%
    \fi}

\iccvfinalcopy %

\def\iccvPaperID{****} %

\ificcvfinal\fi

\begin{document}

\title{A Multidimensional Analysis of Social Biases in Vision Transformers}

\author{Jannik Brinkmann\thanks{\,Corresponding author.}
\hspace{5mm} Paul Swoboda \hspace{5mm} Christian Bartelt\\
University of Mannheim\\
{\tt\small \{jannik.brinkmann, paul.swoboda, christian.bartelt\}@uni-mannheim.de}
}

\maketitle
\ificcvfinal\thispagestyle{empty}\fi

\begin{abstract}
   The embedding spaces of image models have been shown to encode a range of social biases such as racism and sexism.
    Here, we investigate specific factors that contribute to the emergence of these biases in Vision Transformers (ViT).
    Therefore, we measure the impact of training data, model architecture, and training objectives on social biases in the learned representations of ViTs. 
    Our findings indicate that counterfactual augmentation training using diffusion-based image editing can mitigate biases, but does not eliminate them. 
    Moreover, we find that larger models are less biased than smaller models, and that models trained using discriminative objectives are less biased than those trained using generative objectives.
    In addition, we observe inconsistencies in the learned social biases. 
    To our surprise, ViTs can exhibit opposite biases when trained on the same data set using different self-supervised objectives. 
    Our findings give insights into the factors that contribute to the emergence of social biases and suggests that we could achieve substantial fairness improvements based on model design choices. 
\end{abstract}

\section{Introduction}
In recent studies, state-of-the-art self-supervised image models such as SimCLR~\cite{SimCLRv2} and iGPT~\cite{iGPT} have been shown to encode a range of social biases, such as racism and sexism~\cite{iEAT}.
This can lead to representational harm~\cite{BiasIsRepresentationalHarm} and ethical concerns in different socio-technical application scenarios~\cite{tian2020}.
The distributional nature of these models is suspected to be an important factor contributing to the emergence of social biases, as it has been demonstrated that these models tend to encode common co-occurrences of objects associated with social biases~(e.\,g.\,women are more often set in ``home or hotel'' scenes, whereas men are more often depicted in ``industrial and construction'' scenes ~\cite{REVISE}).
Moreover, it has been demonstrated that self-supervised training objectives can impact the distribution of social biases in models that share the same ResNet50~\cite{ResNet50} architecture~\cite{StudySSL}.

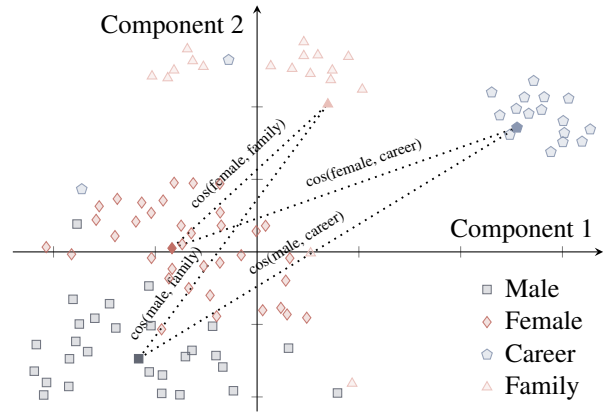
\begin{figure}[t]
  \centering
  \hspace{-0.1cm}\begin{tikzpicture}
\pgfplotsset{%
    width=0.54\textwidth,
    height=0.4\textwidth
}
\begin{axis}[%
    scatter/classes={%
                male={fill=morecontrast-one, draw=morecontrast-one, mark=square*, mark size=1.6},
                female={fill=morecontrast-five, draw=morecontrast-five, mark=diamond*, mark size=2.1},
                career={fill=morecontrast-two, draw=morecontrast-two, mark=pentagon*, mark size=2.1},
                family={fill=morecontrast-six, draw=morecontrast-six, mark=triangle*, mark size=2.1}},
    axis x line=middle,
    axis y line=middle,
    xlabel={Component 1},
    ylabel={Component 2},
    xticklabels={,,},
    yticklabels={,,},
    xtick={-60, -50, -40, -30, -20, -10, 0, 10, 20, 30},
    ytick={-40, -30, -20, -10, 0, 10, 20},
    xmin=-24,
    xmax=34,
    ymin=-22,
    ymax=34,
    legend cell align={left},
    legend style={draw=none, at={(0.78,0.18)},anchor=west},
    y label style={
        yshift=0cm,
        xshift=-2.2cm
    },
    ]
    \addplot+[scatter, only marks,%
        scatter src=explicit symbolic,
        fill opacity=0.2,
        draw opacity=0.9,
        y filter/.expression={y>30 ? NaN : y},
        y filter/.expression={y<-20 ? NaN : y},
        x filter/.expression={x>34 ? NaN : x},
        x filter/.expression={x<-22 ? NaN : x},
    ] table[meta=label] {
        x y label
        -20.966055 -19.732275    male
        -3.369020 -15.874302    male
        -2.347057 -20.100401    male
        -20.687759 -18.005352    male
        7.899418 -19.456175    male
        -18.504026 -18.553825    male
        -7.040870 -14.437516    male
        -27.352581  -3.080496    male
        -13.780993 -24.167542    male
        -14.481702 -25.978800    male
        -21.891991 -16.448561    male
        -4.370187 -10.251241    male
        -23.563824 -18.091354    male
        5.322057 -22.142675    male
        -9.372590 -19.509808    male
        -16.647129 -13.750689    male
        -4.039658 -14.296788    male
        -13.880973 -10.491219    male
        3.088607 -13.273463    male
        -21.691523 -12.779044    male
        -17.466541  -9.946012    male
        -10.596405 -16.550547    male
        -9.297338 -16.978697    male
        -17.691608   3.842010    male
        -10.631347  -4.697917    male
        -10.220538 -20.606758    male
        -17.812305 -12.345944    male
        -15.944539  -9.148539    male
        -6.396942 -13.580544    male
        -18.086470  -6.539324    male
        -26.526165   0.021093    male
        -10.427466 -10.169123    male
        -22.481686 -15.036654    male
        7.168125 -21.535261    male
        -7.585046 -19.700542    male
        6.998865 -20.262009    male
        -15.475343 -24.532146    male
        -18.434168 -14.735961    male
        -2.326978 -19.147949    male
        -14.494772  -7.146076    male
        -8.345151  -2.339775  female
        -11.334577   6.799222  female
        -25.034267   1.272171  female
        -1.560383  -0.450029  female
        -3.654480  -1.486410  female
        -15.813729   4.419388  female
        -3.976140  -8.877108  female
        2.814109  -3.960083  female
        -13.962774   2.178509  female
        -6.273711   9.465919  female
        -27.507685  -0.448254  female
        -5.239226  -1.953042  female
        -7.357404   1.063489  female
        -9.231900   7.087817  female
        -3.660328   9.393873  female
        -18.229712  -0.325154  female
        0.830324   3.602928  female
        4.875933  -9.077130  female
        -6.712290  -4.994956  female
        -9.381516 -10.675551  female
        -23.135590   1.061109  female
        -10.377316  -0.836802  female
        2.948272  -8.624187  female
        -15.512337   6.294619  female
        -20.718025   0.661045  female
        -0.085796   2.866737  female
        -6.467247   2.568186  female
        -26.296255   2.918866  female
        -7.462996   3.616436  female
        0.550371  -8.044477  female
        -13.736269   7.311864  female
        -3.624747   5.439533  female
        -3.523916   3.630016  female
        -4.619592  -5.994466  female
        -8.148617   9.531123  female
        -8.667142  -3.655231  female
        -11.908783   4.179353  female
        -10.399891   5.379241  female
        2.336463  -7.718670  female
        3.211210  -0.868274  female
        38.180088   9.468317  career
        27.172041  23.575382  career
        25.465481  19.733116  career
        -2.798138  26.471350  career
        26.259989  21.310093  career
        34.749104   8.368276  career
        32.729721  16.895639  career
        28.864573  13.773836  career
        27.950069  19.584509  career
        35.551559   8.871632  career
        23.843693  18.829893  career
        30.078037  18.047838  career
        28.204792  21.337122  career
        30.043846  15.138463  career
        23.047461  21.991726  career
        -17.228182   8.648500  career
        26.689478  19.058603  career
        30.605812  21.142704  career
        24.828447  16.140800  career
        31.927116  15.204838  career
        30.184778  16.402523  career
        10.336528  22.407309  family
        -8.163664  25.049746  family
        5.235301  24.568525  family
        -6.853261  27.973137  family
        5.258822  -0.167380  family
        38.701355  11.909453  family
        3.439506  25.265965  family
        4.587441  27.037031  family
        -5.276168  25.545181  family
        7.240111  23.717743  family
        -8.778268  23.977886  family
        7.391323  25.045834  family
        4.588695  28.945616  family
        -10.323744  24.226650  family
        1.344786  25.637463  family
        8.973661  26.921585  family
        40.001801  11.437786  family
        9.337337 -18.143454  family
        5.969910  27.138132  family
        -6.326019  26.331459  family
        39.287560  12.882395  family
        };
        \legend{\hspace{0.1cm}Male, \hspace{0.1cm}Female, \hspace{0.1cm}Career, \hspace{0.1cm}Family}

    \addplot[scatter, only marks,%
        scatter src=explicit symbolic, mark size=5, fill opacity=1,
        draw opacity=1,
    ] table[meta=label] {
        x y label
        -11.635163 -14.730461 male
        -8.359828 0.510296 female
        25.540465 17.142627 career
        6.951096  20.367051 family
        };
    \draw[dotted, line width=0.25mm, draw=black] (-11.635163,-14.730461) -- (25.540465,17.142627) node[midway, above, sloped, font=\fontsize{6pt}{6pt}\selectfont, xshift=-10pt] (TextNode) {cos(male, career)};
    \draw[dotted, line width=0.25mm, draw=black] (-11.635163,-14.730461) -- (6.951096,20.367051) node[midway, above, sloped, font=\fontsize{6pt}{6pt}\selectfont, xshift=-34pt] (TextNode) {cos(male, family)};
    \draw[dotted, line width=0.25mm, draw=black] (-8.359828,0.510296) -- (25.540465,17.142627) node[midway, above, sloped, font=\fontsize{6pt}{6pt}\selectfont, xshift=10pt] (TextNode) {cos(female, career)};
    \draw[dotted, line width=0.25mm, draw=black] (-8.359828,0.510296) -- (6.951096,20.367051) node[midway, above, sloped, font=\fontsize{6pt}{6pt}\selectfont] (TextNode) {cos(female, family)};
\end{axis}
\end{tikzpicture}
  \caption{Gender bias in image embedding from ViTMAE: t-SNE (n=2) reveals that ``female'' is more closely associated with ``family'' rather than ``career'', whereas ``male'' has a comparable association with both attributes.}
  \label{fig:tsne}
  \vspace*{-0.3cm}
\end{figure}

\noindent However, existing work has done little investigation into other factors that contribute to the emergence of social biases in image models. \vspace{-0.1cm}

\paragraph{Contributions} 
Here, we seek to better understand the factors that contribute to the emergence of social biases in image models.
Therefore, we investigate social biases in embedding spaces, which, despite not being observable for end-users, could propagate into downstream tasks during fine-tuning. 
This can help to make informed choices about the model to select for a downstream task, and to develop effective strategies to mitigate social biases.
In detail, the contributions of our work are: 
\begin{itemize}[leftmargin=*]
    \itemsep0em
    \item Training ViTs with counterfactual data augmentation using diffusion-based image editing can reduce social biases, but is not sufficient to eliminate them. 

    \item ViTs trained using discriminative objectives are less biased than those trained using generative objectives.
    
    \item Scaling ViTs can help to mitigate social biases.

    \item ViTs can exhibit opposite biases despite being trained on the same data set, which indicates that biases are not just a result of simple object co-occurrences.
\end{itemize}

\begin{figure*}
\centering
  \begin{subfigure}[t]{0.3\textwidth}
        \picdims[height=0.8in]{0.8in}{0.8in}{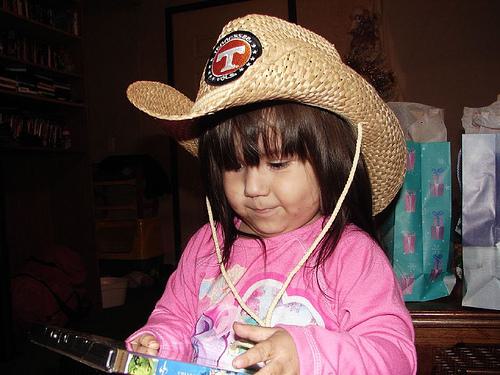}
        \hspace{0.0cm}
        \picdims[height=0.8in]{0.8in}{0.8in}{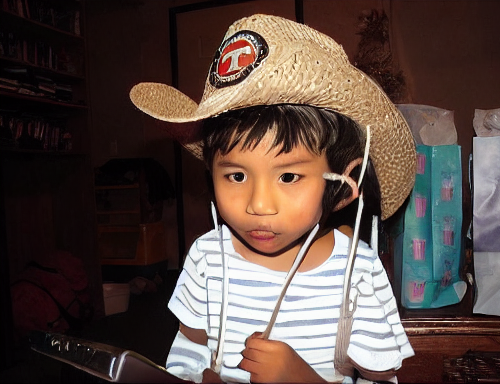}
    \end{subfigure} 
    \begin{subfigure}[t]{0.3\textwidth}
        \picdims[height=0.8in]{0.8in}{0.8in}{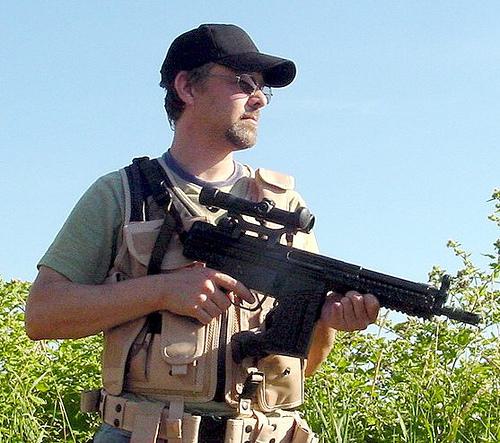}
        \hspace{0.0cm}
        \picdims[height=0.8in]{0.8in}{0.8in}{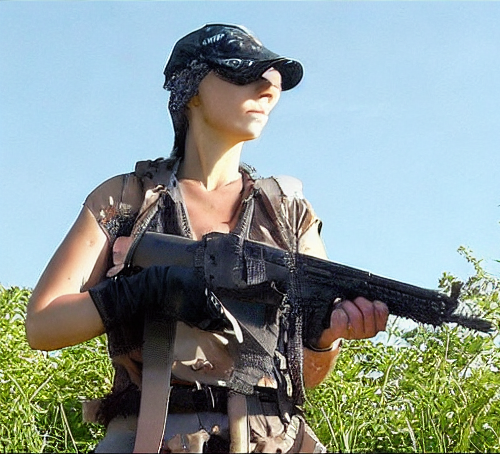}
    \end{subfigure} 
    \begin{subfigure}[t]{0.3\textwidth}
        \picdims[height=0.8in]{0.8in}{0.8in}{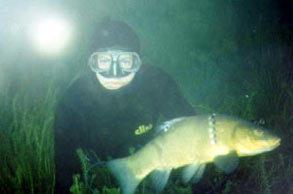}
        \hspace{0.0cm}
        \picdims[width =0.8in]{0.8in}{0.8in}{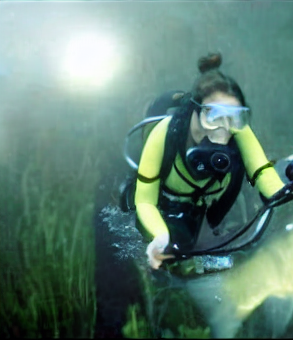}
    \end{subfigure} 
    \vspace{0.2cm}
    \hspace*{-0.8cm}
    \\
    \begin{subfigure}[t]{0.3\textwidth}
        \picdims[height=0.8in]{0.8in}{0.8in}{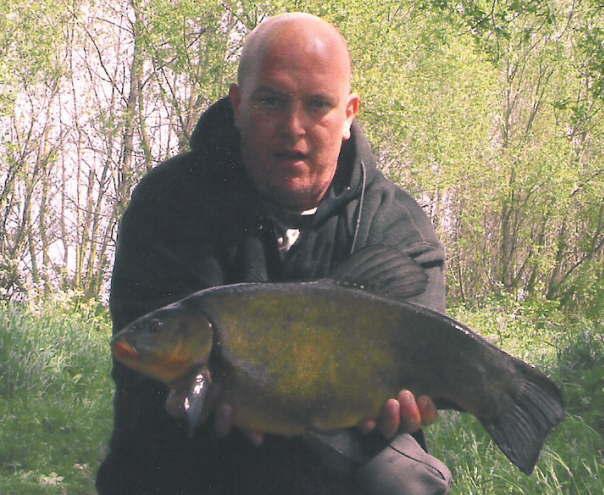}
        \hspace{0.0cm}
        \picdims[height=0.8in]{0.8in}{0.8in}{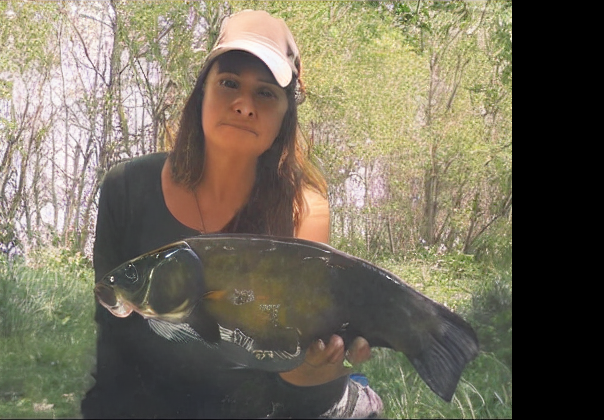}
    \end{subfigure} 
    \begin{subfigure}[t]{0.3\textwidth}
        \picdims[width=0.8in]{0.8in}{0.8in}{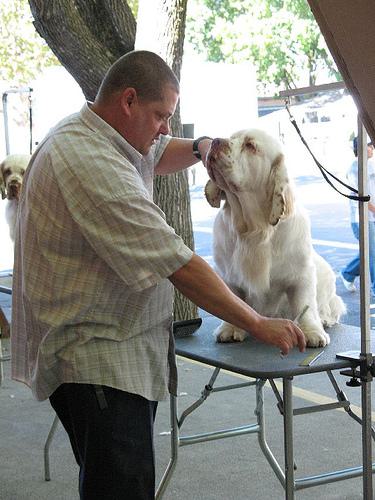}
        \hspace{0.0cm}
        \picdims[width=0.8in]{0.8in}{0.8in}{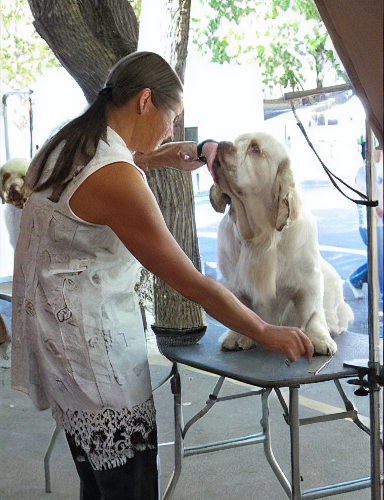}
    \end{subfigure} 
    \begin{subfigure}[t]{0.3\textwidth}
        \picdims[height=0.8in]{0.8in}{0.8in}{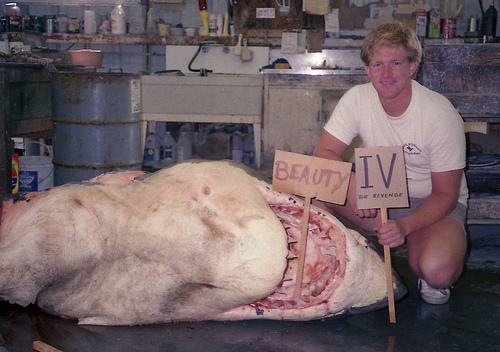}
        \hspace{0.0cm}
        \picdims[height=0.8in]{0.8in}{0.8in}{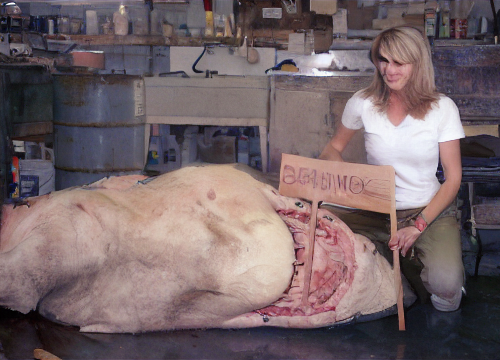}
    \end{subfigure} 
    \vspace{0.2cm}
    \hspace*{-0.8cm}
    \\
    \begin{subfigure}[t]{0.3\textwidth}
        \picdims[width=0.8in]{0.8in}{0.8in}{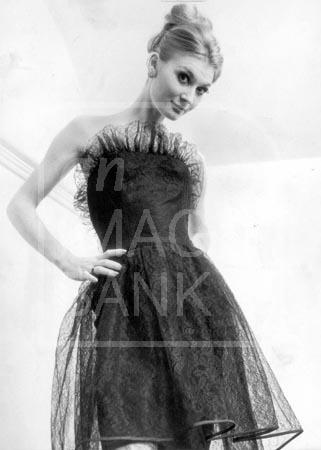}
        \hspace{0.0cm}
        \picdims[width=0.8in]{0.8in}{0.8in}{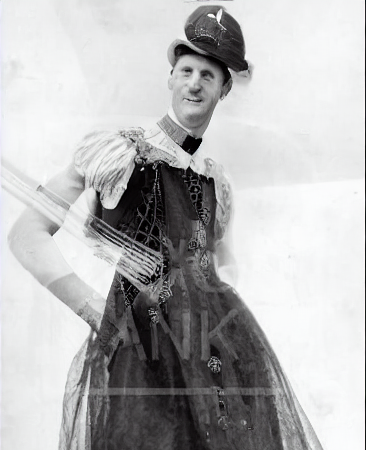}
    \end{subfigure} 
    \begin{subfigure}[t]{0.3\textwidth}
        \picdims[height=0.8in]{0.8in}{0.8in}{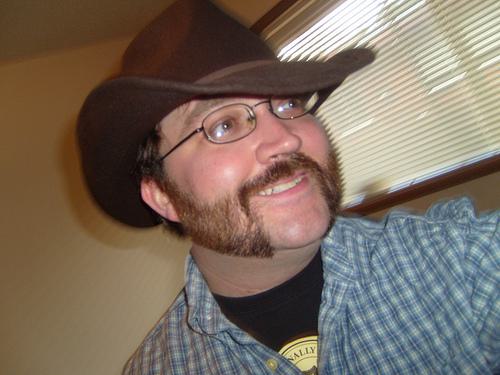}
        \hspace{0.0cm}
        \picdims[height=0.8in]{0.8in}{0.8in}{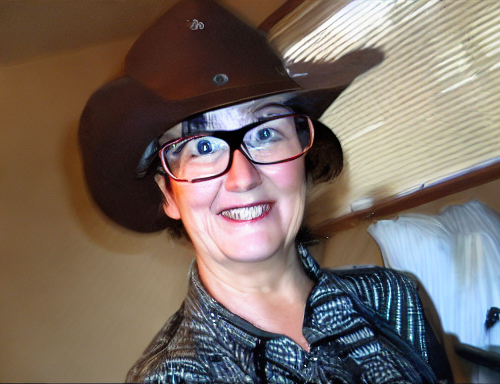}
    \end{subfigure} 
    \begin{subfigure}[t]{0.3\textwidth}
        \picdims[width=0.8in]{0.8in}{0.8in}{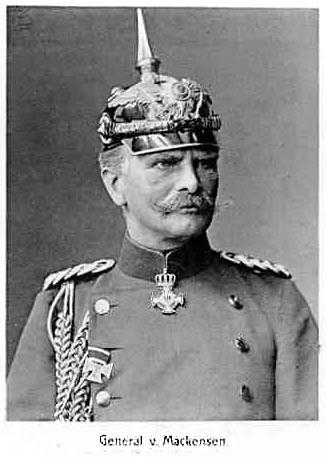}
        \hspace{0.0cm}
        \picdims[width=0.8in]{0.8in}{0.8in}{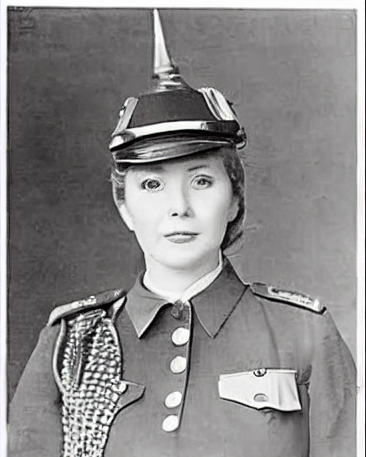}
    \end{subfigure} 
    \vspace{0.2cm}
    \hspace*{-0.8cm}
    \\
    \begin{subfigure}[t]{0.3\textwidth}
        \picdims[height=0.8in]{0.8in}{0.8in}{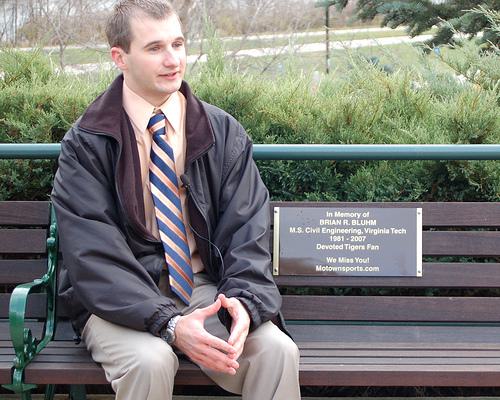}
        \hspace{0.0cm}
        \picdims[height=0.8in]{0.8in}{0.8in}{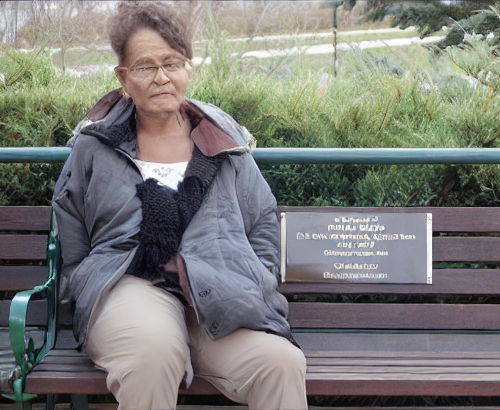}
    \end{subfigure} 
    \begin{subfigure}[t]{0.3\textwidth}
        \picdims[width=0.8in]{0.8in}{0.8in}{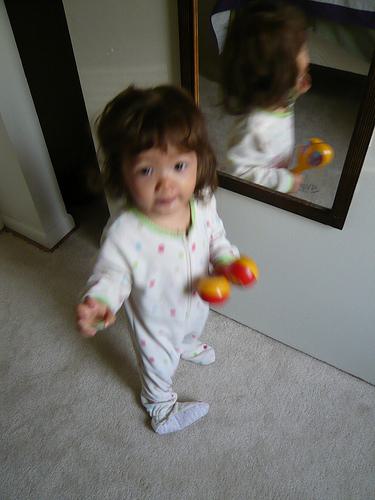}
        \hspace{0.0cm}
        \picdims[width=0.8in]{0.8in}{0.8in}{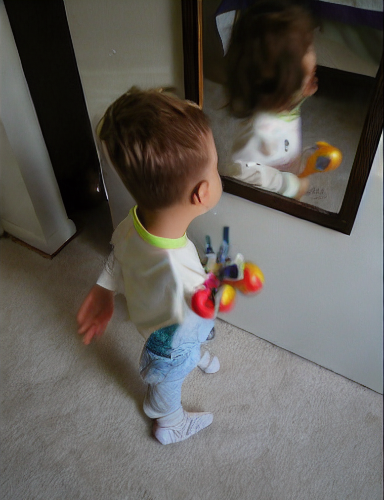}
    \end{subfigure} 
    \begin{subfigure}[t]{0.3\textwidth}
        \picdims[height=0.8in]{0.8in}{0.8in}{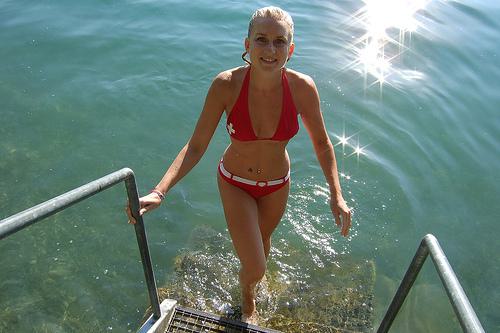}
        \hspace{0.0cm}
        \picdims[height=0.8in]{0.8in}{0.8in}{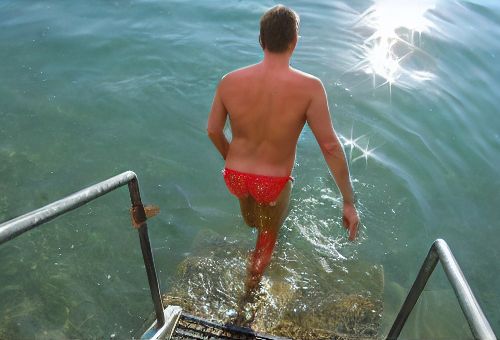}
    \end{subfigure} 
    \vspace{0.2cm}
    \hspace*{-0.8cm}
    \\
    \begin{subfigure}[t]{0.3\textwidth}
        \picdims[width=0.8in]{0.8in}{0.8in}{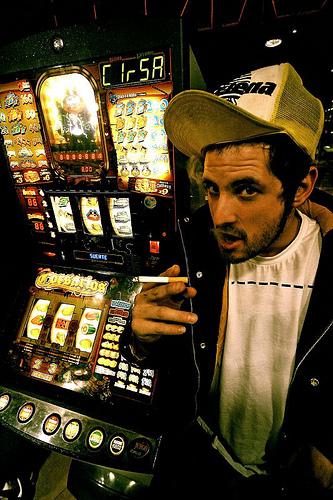}
        \hspace{0.0cm}
        \picdims[width=0.8in]{0.8in}{0.8in}{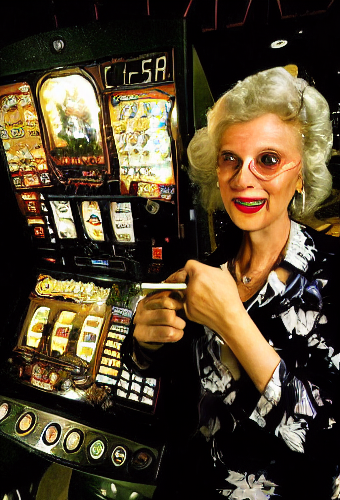}
    \end{subfigure} 
    \begin{subfigure}[t]{0.3\textwidth}
        \picdims[height=0.8in]{0.8in}{0.8in}{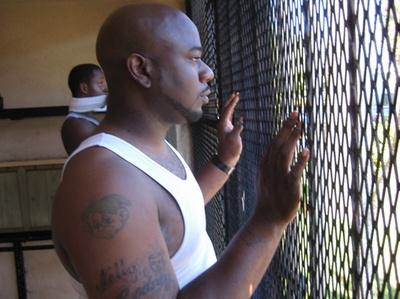}
        \hspace{0.0cm}
        \picdims[height=0.8in]{0.8in}{0.8in}{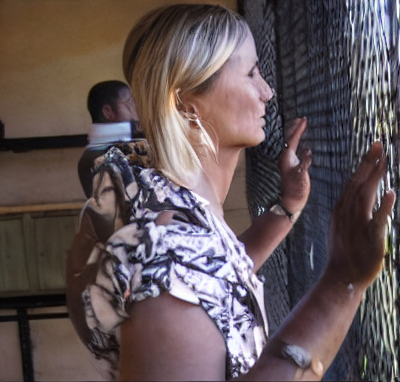}
    \end{subfigure} 
    \begin{subfigure}[t]{0.3\textwidth}
        \picdims[height=0.8in]{0.8in}{0.8in}{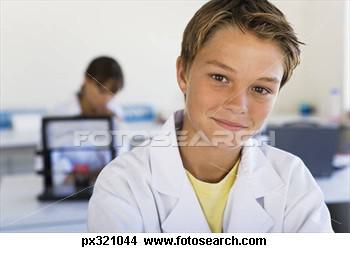}
        \hspace{0.0cm}
        \picdims[height=0.8in]{0.8in}{0.8in}{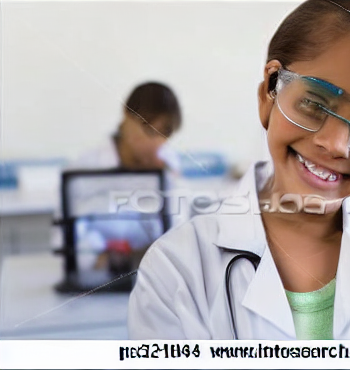}
    \end{subfigure}
    \hspace*{-0.8cm}
  \caption{\textbf{Selected counterfactual images on ImageNet.} In each case, we show the original image (left), and the generated counterfactual image (right).}
  \label{fig:counterfactual-examples}
\end{figure*}

\section{Related Work}
\paragraph{Self-Supervised Learning of ViTs}
Self-supervised approaches have emerged as the standard for training large machine learning models since they don't require labeled data and learn representations that generalize well across different downstream tasks~\cite{GPT}.
Transformer models~\cite{Transformers}, which were designed as sequence-to-sequence models for natural language translation, have been adapted to computer vision~\cite{ViT}.
Self-supervised learning techniques applied to ViTs can be classified into discriminative (or joint-embedding) methods and generative (or reconstruction-based) methods~\cite{shekhar2023objectives}.
Discriminative methods encourage similarity among representations from diverse augmentations of a given input image, while generative methods utilize a reconstruction loss that does not rely on augmentations. 
Instead it uses a decoder to reconstruct the original image given a masked image. 
Both methods have demonstrated strong empirical results on downstream tasks~\cite{DINO, iGPT, ViTMoCo, ViTMAE}.

\paragraph{Social Biases in Image Embeddings}
The embeddings of self-supervised image models have been shown to encode a range of human-like social biases~\cite{iEAT}.
However, the analysis was confined to SimCLR~\cite{SimCLRv2} and iGPT~\cite{iGPT} as embedding models. 
Therefore, Sirotkin~\etal~\cite{StudySSL} built on this work to examine the distribution of social biases in image models that were trained using a range of self-supervised objectives, such as geometric, cluster-based, and contrastive methods. 
The authors discovered that models trained with contrastive methods exhibit the largest number of social biases, and that the distribution of biases differs depending on the studied embedding layer.
However, their analysis focused only on training objectives and the number of social biases without considering the direction of the bias, constraining the interpretability of their findings.
In addition, their investigation was conducted on models using a ResNet50~\cite{ResNet50} architecture, excluding ViTs which are considered the standard for transfer learning~\cite{VisionTransformersSurvey}.

\paragraph{Bias Mitigation Methods}
The approaches to mitigate biases can be distinguished into methods that manipulate the training data and methods that adjust the training procedure~\cite{SurveyBiasInML}. 
To mitigate biases during training, existing work suggests, amongst others, adversarial learning~\cite{AdversarialLearningVisualRecognition}, training separate models for each attribute~\cite{SurveyFairnessInCV}, or incorporating regularization terms~\cite{RegularizationForBiasMitigation, Jung_2021_CVPR}.
In contrast, the methods to mitigate biases in the training data aim to generate unbiased data sets that are balanced~\cite{FairFace} or do not include information about the bias dimension~\cite{GenderArtifactsInVisualDatasets}. 
One approach to mitigate biases in the training data is Counterfactual Data Augmentation~(CDA)~\cite{CDA}. 
This method entails generating training instances that contradict the observed biases.
There are different variations of CDA: 1-sided CDA, which use just the counterfactuals during an additional pre-training phase, and 2-sided CDA, which uses both counterfactuals and the original training data.
While 1-sided CDA has a more substantial impact on biases, it can lead to over-correction~\cite{Webster}. 
In existing work, CDA has been used to mitigate different types of biases in language models~\cite{LauscherAdapterDebiasing}, operating on a set of term pairs, such as ``man'' and ``woman''.
However, generating counterfactual training instances from images is non-trivial. 
To address this, conditional generative adversarial networks have been used to generate unbiased training data with balanced protected attributes~\cite{Ramaswamy_2021_CVPR, GANForUnbiasedDataset}. 
Therefore, the authors generate multiple synthetic images for each training image, maintaining the target attribute score but reversing the expression score on the protected attribute. 
These approaches have demonstrated to be effective at mitigating bias on selected dimensions, but do not eliminate them. 
In addition, existing methods focus on downstream tasks and no research has been conducted on debiasing pre-trained image models used as backbones for transfer learning. 

\section{Background}
\paragraph{iEAT}
The Image Embedding Association Test (iEAT) quantifies social biases in image embeddings based on semantic similarities~\cite{iEAT}. 
It compares the differential association of image embeddings of selected target concepts (such as ``male'' and ``female'') and attributes (such as ``science'' and ``liberal arts''), and tests the null-hypothesis of equal similarities of the target concepts and attributes.
Hence, a rejection suggests that one target concept is more associated with one attribute than the other (such as ``male'' is more associated with ``science'' or ``female'' is more associated with ``liberal arts''). 
To test the null-hypothesis, it formulates a test statistic that compares target concepts X and Y with attributes A and B, defined as:
$$s(X, Y, A, B) = \sum_{x \in X}^{} s(x, A, B) - \sum_{y \in Y}^{} s(y, A, B)$$
where $s(w, A, B)$ is the differential association of a target concept with the attributes, measured using the cosine similarities of their embeddings:
$$s(w, A, B) = \mu({cos(w, a)}_{a \in A}) - \mu({cos(w, b)}_{b \in B})$$
where $\mu$ is the mean. 
The statistical significance is determined using a permutation test, contrasting the score $s(X, Y, A, B)$ with the scores $s(X_i, Y_i, A, B)$, where $X_i$ and $Y_i$ are all equal-sized partitions of the set $X\,\cup\,Y$: 
\begin{equation}
    p_t = Pr[s(X_i, Y_i, A, B) > s(X, Y, A, B)]
    \label{equ:test-statistic}
\end{equation}
The effect size $d$ quantifies the bias magnitude, computed as the normalized separation of the association distributions:
\begin{equation}
    d = \frac{\mu({s(x, A, B)}_{x \in X}) -\mu({s(y, A, B)}_{y \in Y})}{\sigma({s(t, A, B)}_{t\ \in\, X \cup Y})}
    \label{equ:effect-size}
\end{equation}
where $\mu$ is the mean and $\sigma$ is the standard deviation. 
Here, the distance from zero indicates the bias magnitude, such that an effect size equaling zero implies the absence of bias.
Moreover, the effect size indicates the direction of the bias, such that a negative effect size suggests that the differential association of Y with A and B is more pronounced, whereas a positive effect size implies the opposite scenario.

The iEAT framework introduces a collection of 15 association tests designed to measure human-like social biases~(see~Table~\ref{tab:ieat-specs}). 
These tests offer a valuable baseline to assess the presence and intensity of certain social biases within image embeddings. 
However, it it important to recognize that these are not an exhaustive list of all possible biases. 
These biases were selected due to their recurrence in related literature and societal implications. 
However, there might be other biases not captured in this selection, such as political biases. 
Nonetheless, these tests remain an instrumental foundation to assess the existence and magnitude of social biases in image embeddings.

\paragraph{Embedding Layer}
The selection of an embedding layer is crucial to extract features that contain high-quality, general-purpose information about the objects in an image. 
It has been demonstrated that in ViTs trained with supervised methods, the model depth tends to correlate with the quality of the embeddings, with the highest-quality embeddings being in the second-to-last layer~\cite{EmbeddingsInSupervisedSettings}. 
In contrast, ViTs trained with SSL methods have been found to generate the most useful embeddings at a layer in the middle of the model~\cite{BEiT, iGPT}. 
Therefore, the selection of an embedding layer depends on the training approach and the specific model. 
Here, for each model, we choose the layer that has been reported to be optimal in linear evaluations.

\begin{table*}[!b]
    \label{tab:ieat-specifications}
    \small
    \setlength\extrarowheight{-2pt}
    \begin{tabularx}{\textwidth}{l X X X X}
        \toprule
            \textsc{Test} & \textsc{Target A} & \textsc{Target B} & \textsc{Attribute X} & \textsc{Attribute Y} \\ 
        \midrule
            T1 & Young & Old & Pleasant & Unpleasant \\
            T2 & Other & Arab-Muslim & Pleasant & Unpleasant \\
            T3 & European American & Asian American & American & Foreign \\
            T4 & Disabled & Not-Disabled & Pleasant & Unpleasant \\
            T5 & Male & Female & Career & Family \\
            T6 & Male & Female & Science & Liberal Arts \\
            T7 & Flower & Insect & Pleasant & Unpleasant \\
            T8 & European American & Native American & Pleasant & Unpleasant \\
            T9 & European American & African American & Pleasant & Unpleasant \\
            T10 & Christianity & Judaism & Pleasant & Unpleasant \\
            T11 & Gay & Straight & Pleasant & Unpleasant \\
            T12 & Light Skin & Dark Skin & Pleasant & Unpleasant \\
            T13 & White & Black & Tool & Weapon \\
            T14 & White & Black & Tool & Weapon (Modern) \\
            T15 & Thin & Fat & Pleasant & Unpleasant \\
        \bottomrule
    \end{tabularx}
    \centering
    \caption{Image Embedding Association Tests\label{tab:ieat-specs}}
\end{table*}

\section{Experiments and Results}
Here, we describe and discuss our experiments to investigate factors that contribute to the emergence of social biases in the embedding spaces of ViTs. 
Therefore, we assess bias mitigation methods along multiple dimensions: 
\begin{itemize}[leftmargin=*]
    \itemsep0.8em
    \item Training data: We investigate counterfactual augmentation training using diffusion-based image editing and find that it can reduce social biases in ViTs, but is not sufficient to eliminate them (Section~\ref{cha:exp-training-data}). 
    \item Training objectives: We assess the impact of training objectives, and find that ViTs trained using discriminative objectives are less biased than those trained using generative objectives (Section~\ref{cha:exp-objectives}).
    \item Model architecture: We evaluate the impact of different architectural choices and find that social biases decrease as model size and input resolution increase, but observe no systematic effect for patch size (Section~\ref{cha:exp-architectural-choices}).
\end{itemize}

\vspace{-0.1cm}
\subsection{Impact of Training Data}\label{cha:exp-training-data}
The emergence of social biases in self-supervised image models is often suggested to be a result of object co-occurrences in images (\eg women are more often set in ``home or hotel'' scenes, whereas men are more often depicted in ``industrial and construction'' scenes~\cite{REVISE}).
However, little research has been conducted on the effect of modifications of the training data on social biases in pre-trained image models.
Therefore, we investigate the debiasing effect of counterfactual data on gender bias as an example. 
Our findings suggest that it can reduce social biases both during pre-training and fine-tuning, although it does not eliminate them and can come at a cost of a slight reduction in downstream performance. 
Moreover, we observe differences in the responsiveness to the counterfactual data, suggesting that its effectiveness is model-specific. 

\paragraph{Models} 
In our experiments, we use BEiT~\cite{BEiT}, ViT-MoCo~\cite{ViTMoCo} and ViT-MAE~\cite{ViTMAE}, which use a standard Transformer as the backbone network (12 layers, 12 attention heads, 768 hidden size). The implementation and model weights were made available using HuggingFace's Transformers~\cite{Transformers} and Timm~\cite{rw2019timm}.

\paragraph{Counterfactual Data Augmentation} 
To investigate the impact of training data, we examine to what extent counterfactual data augmentation can mitigate social biases in ViTs. 
In our experiments, we combine the approach to counterfactual data augmentation used in natural language processing with diffusion-based image editing. 
Therefore, we leverage a large-scale text-to-image diffusion model~\cite{StableDiffusion} as a foundation, to capitalize on the benefits of pre-training on a sizable and generic corpus. 
For each image, we generate a textual description using BLIP~\cite{BLIP} and CLIP~\cite{CLIP}.
Then, we use a set of term pairs (\eg ``man'', ``woman'') to substitute target words in the generated caption.
For our purposes, we adopt the set of gender term pairs of Zhao \etal~\cite{zhang2018mitigating}.
To generate counterfactual images, we use diffusion-based semantic image editing with mask guidance ~\cite{DiffEdit}.
To this end, we use CLIPSeg~\cite{CIDAS} to mask the target words (\eg ``man'') in the image and use Stable Diffusion~\cite{SDInpainting} to inpaint the masked image section, conditioned on the modified captions~(see Figure~\ref{fig:counterfactual-examples}).

Here, we adopt the ImageNet ILSVRC 2012 dataset~(ImageNet-1K)~\cite{ILSVRC15} as our benchmark to assess the effectiveness of the generated data, as it is one of the most studied benchmarks for which there is an extensive literature on architecture and training procedures.
ImageNet-1K contains 1.28 million images, from which we generate an additional 159,393 counterfactual images.

\paragraph{Counterfactual Training}
To evaluate the debiasing effect of counterfactual data, we follow Webster \etal~\cite{Webster} and continue the training of the models from a pre-trained checkpoint using the counterfactual images (1-sided CDA). 
To this end, we adopt the standard contrastive learning objective for ViT-MoCo~\cite{ViTMoCo} and masked image modeling training objective for BEiT and ViT-MAE with a masking ratio of 40 \%~\cite{BEiT} and 75~\%~\cite{ViTMAE}, respectively. 
Then, we train each model using Adam~\cite{Adam} with a batch size of 128 and learning rate 1.5e-4 for a single epoch to avoid over-correction~\cite{Webster}. 
The results are depicted in Table~\ref{tab:cda-fine-tuning}.

\begin{table}[!ht]
    \centering
    \begin{tabular}{l | rr | rr}
    \toprule
    & \multicolumn{2}{c|}{\textsc{ Baseline }} & \multicolumn{2}{c}{\textsc{CDA}} \\
    \textsc{Model} & \textsc{Bias} & \textsc{Cifar10} & \textsc{Bias} & \textsc{Cifar10} \\
    \midrule
    \textsc{BEiT} & 0.65 & \textbf{87.5} & \textbf{0.45} & 84.8 \\ 
    \textsc{ViT-MoCo} & 1.41 & \textbf{95.1} & \textbf{1.39} & \textbf{95.1} \\
    \textsc{ViT-MAE} & \textbf{0.59} & \textbf{89.6} & 0.64 & \textbf{89.6} \\
    \bottomrule
    \end{tabular}
    \caption{
    iEAT effect size (see Equation~\ref{equ:effect-size}) and linear evaluation performance on CIFAR10 of different models before (Baseline) and after (CDA) debiasing using a single pre-training epoch on counterfactual data. 
    \textbf{We find that counterfactual data augmentation can reduce social biases, but its effect is model-specific and can come with a reduction in representation quality.}
    }
    \label{tab:cda-fine-tuning}
\end{table}

In addition to the gender bias, we report the linear evaluation performance on CIFAR10~\cite{CIFAR} as a measure of representation quality.
We observe that it does reduce gender bias on BEiT and ViT-MoCo but comes with a slight reduction in representation quality for BEiT.
However, a similar effect has been observed in alternative debiasing methods before and is not specific to out setting~\cite{rakabsaz, zerveas}.
In contrast, we observe the opposite effect on ViT-MAE, where it comes with a small increase in gender bias. 
This implies that there are differences in the responsiveness to the counterfactual data, suggesting that the effectiveness of this technique might be model-specific.
We hypothesis that this is a result of the training objectives, which could influence how the models learn from the counterfactual data.
In addition, we conjecture that the counterfactual data could interact differently with pre-trained checkpoints, which could carry certain biases leading to varying debiasing effects. 

\begin{table}[!ht]
    \centering
    \begin{tabular}{l | rr | rr}
    \toprule
    & \multicolumn{2}{c|}{\textsc{ Baseline }} & \multicolumn{2}{c}{\textsc{CDA}} \\
    \textsc{Model} & \textsc{Bias} & \textsc{Cifar10} & \textsc{Bias} & \textsc{Cifar10} \\
    \midrule
    \textsc{ViT-MoCo} & 1.25 & 90.4 & \textbf{1.04} & \textbf{90.9} \\
    \textsc{ViT-MAE} & \textbf{0.50} & \textbf{82.9} & 0.55 & 71.2 \\
    \bottomrule
    \end{tabular}
    \caption{
    iEAT effect size (see Equation~\ref{equ:effect-size}) and linear evaluation performance on CIFAR10 of different models pre-trained from scratch on ImageNet-1k (Baseline), and both ImageNet-1k and the counterfactual data (CDA).
    \textbf{We again observe a decrease in gender bias on MoCo-v3 and increase on ViTMAE.
    This implies that the observed effects are not a result of the pre-trained checkpoint.}
    }
    \label{tab:cda-pre-training}
\end{table}

To evaluate whether the observed effects on ViT-MoCo and ViT-MAE are a result of their pre-trained checkpoints, we train them from scratch on ImageNet-1k and our counterfactual data (2-sided CDA). 
The results are illustrated in Table~\ref{tab:cda-pre-training}.
We again observe a decrease in gender bias on ViT-MoCo, and a similar increase in gender bias on ViT-MAE. 
This implies that observed effects are not a result of the pre-trained checkpoint, and that other factors influence the debiasing effect, such as model architecture differences.
These findings highlight the nuanced effect of training data on social biases, demanding tailored approaches for different architectures and training approaches.
Thus, we anticipate the need for more principled approaches that eliminate undesirable model behavior, potentially bypassing the use of counterfactual data and instead using post-hoc interventions to eliminate biases directly.

\begin{table*}[!t]
    \small
    \setlength{\tabcolsep}{0.22pt}
    \begin{tabular*}{\textwidth}{l*{15}{x{10mm}}}
        \toprule
            \textsc{Models} & \textsc{T1} & \textsc{T2} & \textsc{T3} & \textsc{T4} & \textsc{T5} & \textsc{T6} & \textsc{T7} & \textsc{T8} & \textsc{T9} & \textsc{T10} & \textsc{T11} & \textsc{T12} & \textsc{T13} & \textsc{T14} & \textsc{T15} \tn
            
        \midrule

            \multicolumn{15}{l}{\textsc{Discriminative Models}}\tn
            
            \textsc{ViT-DINO-B} & \boldsymbol{$\textcolor{white}{-}0.99$} & \boldsymbol{$\textcolor{white}{-}1.20$} & $-0.86$ & $\textcolor{white}{-}0.88$ & \boldsymbol{$\textcolor{white}{-}0.38$} & $\textcolor{white}{-}0.01$ & $-0.12$ & \boldsymbol{$\textcolor{white}{-}0.84$} & $\textcolor{white}{-}0.49$ & $\textcolor{white}{-}0.22$ & $-0.08$ & $-0.13$ & $-0.88$ & $-0.77$ & \boldsymbol{$\textcolor{white}{-}1.24$} \tn
            
            \textsc{ViT-MoCo-B} & $-0.15$ & \boldsymbol{$\textcolor{white}{-}1.02$} & $-0.75$ & $-0.29$ & \boldsymbol{$\textcolor{white}{-}1.41$} & $\textcolor{white}{-}0.13$ & \boldsymbol{$\textcolor{white}{-}1.68$} & $-0.66$ & \boldsymbol{$\textcolor{white}{-}1.10$} & $\textcolor{white}{-}0.46$ & $-0.24$ & $-0.11$ & $\textcolor{white}{-}0.77$ & $\textcolor{white}{-}0.14$ & $\textcolor{white}{-}0.64$ \tn

            \textsc{ViT-MSN-B}~\cite{ViTMSN} & $\textcolor{white}{-}0.93$ & \boldsymbol{$\textcolor{white}{-}1.24$} & $\textcolor{white}{-}0.33$ & $\textcolor{white}{-}0.93$ & $\textcolor{white}{-}0.14$ & $-0.31$ & $\textcolor{white}{-}0.10$ & $\textcolor{white}{-}0.60$ & $-0.78$ & $\textcolor{white}{-}0.54$ & $-0.28$ & \boldsymbol{$-1.09$} & $\textcolor{white}{-}0.18$ & $-0.08$ & \boldsymbol{$\textcolor{white}{-}1.64$} \tn

        \midrule

            \multicolumn{15}{l}{\textsc{Generative Models}} \tn
            
            \textsc{BEiT-B} & $\textcolor{white}{-}0.18$ & $\boldsymbol{\textcolor{white}{-}0.82}$ & $\textcolor{white}{-}0.02$ & $\textcolor{white}{-}0.53$ & \boldsymbol{$\textcolor{white}{-}0.65$} & $-0.09$ & \boldsymbol{$-1.02$} & $\textcolor{white}{-}0.28$ & \boldsymbol{$\textcolor{white}{-}1.28$} & $\textcolor{white}{-}0.09$ & $\textcolor{white}{-}0.26$ & \boldsymbol{$\textcolor{white}{-}1.14$} & $\boldsymbol{-1.58}$ & $\textcolor{white}{-}0.56$ & \boldsymbol{$\textcolor{white}{-}1.72$} \tn
            
            \textsc{iGPT-S} & $\textcolor{white}{-}0.66$ & \boldsymbol{$\textcolor{white}{-}0.84$} & \boldsymbol{$-1.02$} & $\textcolor{white}{-}0.75$ & $\textcolor{white}{-}0.22$ & $\textcolor{white}{-}0.16$ & \boldsymbol{$-0.55$} & \boldsymbol{$-1.32$} & $\textcolor{white}{-}0.54$ & $\textcolor{white}{-}0.28$ & $\textcolor{white}{-}0.29$ & \boldsymbol{$\textcolor{white}{-}1.31$} & \boldsymbol{$-1.11$} & $\textcolor{white}{-}0.89$ & \boldsymbol{$\textcolor{white}{-}1.69$} \tn
            
            \textsc{ViT-MAE-B} & $\textcolor{white}{-}0.11$ & \boldsymbol{$\textcolor{white}{-}0.55$} & $-0.29$ & $-0.35$ & \boldsymbol{$\textcolor{white}{-}0.59$} & $\textcolor{white}{-}0.08$ & \boldsymbol{$-1.15$} & \boldsymbol{$-1.15$} & $-0.81$ & $\textcolor{white}{-}0.34$ & $\textcolor{white}{-}0.29$ & \boldsymbol{$\textcolor{white}{-}0.96$} & \boldsymbol{$-1.30$} & \boldsymbol{$-1.31$} & \boldsymbol{$\textcolor{white}{-}1.75$} \tn

        \bottomrule
    \end{tabular*}
    \centering
    \caption{iEAT effect sizes (see Equation~\ref{equ:effect-size}) for a range of association tests (see Table~\ref{tab:ieat-specs}) using different embedding models. The models were trained on ImageNet-21k using self-supervised methods, with the exception of ViT-MoCo which was trained on ImageNet-1k. The effect sizes indicate the magnitude and direction of the bias, and are written in bold if the effect is significant at $p_t = 0.05$. \textbf{ViTs trained using different self-supervised objectives can exhibit opposite social biases, despite being trained on the same dataset.}
    \label{tab:experiment-models}}
    \vspace*{-0.4cm}
\end{table*}
\subsection{Impact of Training Objectives}\label{cha:exp-objectives}

ResNet50~\cite{ResNet50} models, when trained using different self-supervised objectives exhibit a different number of social biases~\cite{StudySSL}.
Therefore, we investigate the effect of training objectives on biases in ViTs across a range of different self-supervised methodologies: discriminative and generative models.
Our findings indicate that ViTs trained with discriminative learning objective are less biased than those trained using generative objectives. 
Moreover, we observe that models trained on the same dataset using different objectives can exhibit opposite biases, which highlights the importance of training objectives as an important factor in the emergence of social biases in embedding spaces. 

\begin{figure}[!b]
  \centering
  \small
  \begin{tikzpicture}
\pgfplotsset{
    scaled x ticks=false,
    width=0.5\textwidth,
    height=0.35\textwidth
}
\begin{axis}[
    axis x line=middle,
    axis y line=middle,
    xlabel={$p_t$},
    ylabel={Number of Social Biases},
    ytick={0, 4, 8},
    yticklabels={0, 4, 8},
    xtick={0, 0.02, 0.04, 0.06, 0.08, 0.1},
    xticklabels={0, 0.02, 0.04, 0.06, 0.08, 0.1},
    xtick style = {draw=none, font=\small,},
    ytick style={font=\small},
    legend style={draw=none, fill=none, at={(1.02, 0.34)}},
    legend cell align={left},
    y label style={
        rotate=90,
        yshift=1.05cm,
        xshift=-1.72cm
    },
    x label style={
        yshift=-1cm,
        xshift=-1pt
    },
    legend columns=2,
    xmax=0.105,
    ymin=0,
    ymax=10,
    reverse legend,
    y axis line style = {opacity=0},
]

    \addplot[dashed, morecontrast-four, mark=asterisk, mark size=2.2, mark options={solid}, line width=0.5mm] coordinates
     {(0.0001, 0)(0.0002, 1)(0.0026, 1)(0.0027, 2)(0.0174, 2)(0.0175, 3)(0.0551, 3)(0.055200000000000006, 4)(0.0811, 4)(0.08120000000000001, 5)(0.08330000000000001, 5)(0.0834, 6)(0.1, 6)};
     \addlegendentry{MSN}

     \addplot[dashed, morecontrast-five, mark=square, mark size=1.6, mark options={solid}, line width=0.5mm] coordinates
     {(0.0001, 0)(0.0002, 2)(0.0061, 2)(0.006200000000000001, 3)(0.0183, 3)(0.0184, 4)(0.08370000000000001, 4)(0.08380000000000001, 5)(0.0887, 5)(0.0888, 6)(0.09730000000000001, 6)(0.09740000000000001, 7)(0.1, 7)};
     \addlegendentry{MoCo} 

         \addplot[dashed, morecontrast-six, mark=triangle, mark size=1.8, mark options={solid}, line width=0.5mm] coordinates
     {(0.0001, 0)(0.0012000000000000001, 0)(0.0013000000000000002, 1)(0.0022, 1)(0.0023, 2)(0.04050000000000001, 2)(0.040600000000000004, 3)(0.0454, 3)(0.045500000000000006, 4)(0.0478, 4)(0.047900000000000005, 5)(0.0627, 5)(0.06280000000000001, 6)(0.067, 6)(0.0671, 7)(0.0995, 7)(0.09960000000000001, 8)(0.1, 8)};
     \addlegendentry{DINO}

     \addplot[morecontrast-three, line width=0.5mm, mark=diamond, mark size=2] coordinates
     {(0.0001, 0)(0.0002, 1)(0.0015, 1)(0.0016, 2)(0.0037, 2)(0.0038, 3)(0.0077, 3)(0.0078000000000000005, 4)(0.028, 4)(0.0281, 5)(0.0292, 5)(0.0293, 6)(0.0357, 6)(0.035800000000000005, 7)(0.06380000000000001, 7)(0.06390000000000001, 8)(0.1, 8)};
     \addlegendentry{iGPT}

     \addplot[morecontrast-two, line width=0.5mm, mark=pentagon, mark size=2] coordinates
     {(0.0001, 2)(0.0002, 3)(0.0008, 3)(0.0009000000000000001, 4)(0.0108, 4)(0.0109, 5)(0.013900000000000001, 5)(0.014, 6)(0.0313, 6)(0.031400000000000004, 7)(0.1, 7)};
     \addlegendentry{BEiT}

     \addplot[morecontrast-one, line width=0.5mm, mark=Mercedes star flipped, mark size=2.2] coordinates
     {(0.0001, 1)(0.0002, 2)(0.0008, 2)(0.0009000000000000001, 3)(0.0034, 3)(0.0035, 4)(0.007200000000000001, 4)(0.007300000000000001, 5)(0.0097, 5)(0.0098, 7)(0.032600000000000004, 7)(0.03270000000000001, 8)(0.0811, 8)(0.08120000000000001, 9)(0.1, 9)};
     \addlegendentry{MAE}

\end{axis}
\end{tikzpicture}
  \vspace*{-0.6cm}
  \caption{The number of biases detected in embedding spaces of ViTs for different values of $p_t$ (see Equation~\ref{equ:test-statistic}). \textbf{ViTs trained using discriminative objectives are less biased than those trained using generative objectives.}\label{fig:social-bias-distribution}}
\end{figure}
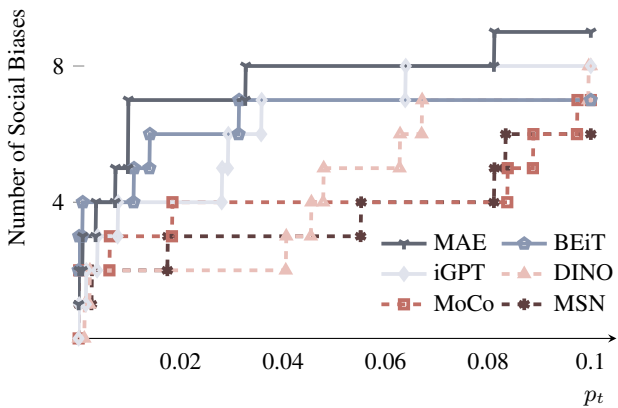

\paragraph{Discriminative and Generative Objectives}
We investigate the distribution of social biases in ViTs trained on ImageNet-21k using different self-supervised objectives.
To this end, we follow Sirotkin~\etal~\cite{StudySSL} and count the number of significant social biases across different values of $p_t$ (see Equation~\ref{equ:test-statistic}) in the range of~$[10^{-4}, 10^{-1}]$, where lower values of $p_t$ correspond to higher statistical significance of the social biases.
The results of this analysis are illustrated in Figure~\ref{fig:social-bias-distribution}.
Our findings indicate that, on average, ViTs trained using discriminative objectives exhibit fewer biases than those trained using generative objectives.
This effect remains consistent across all threshold values, which highlights the robustness of our findings.
We conjecture that this stems from the inherent characteristics of models trained using generative objectives, which encourage the model to reconstruct images that match the statistical patterns in the training data, capture underlying structure and dependencies within the data.
Thus, if the training data is biased towards specific demographics, objects, or scenes, the model could unintentionally learn and perpetuate those biases in its representations.
In contrast, discriminative learning objectives encourage representations that maximize view invariance between samples from the same image~\cite{shekhar2023objectives}.
This encourages the model to learn and prioritize fundamental visual features that are less influenced by social biases or external factors.

\paragraph{Opposite Biases despite same Training Data} 
The analysis of the number of significant biases fails to capture their direction. 
To address this, we contrast the effect sizes in Table~\ref{tab:experiment-models}. 
To our surprise, we find that ViTs can exhibit opposite social biases, despite being trained on the same dataset, \eg ViTMAE exhibits a tendency to perceive Native Americans as less pleasant than European Americans, while ViT-MoCo~\cite{ViTMoCo} exhibits the inverse association.
However, we also find that all models reinforce a handful of consistent social biases irrespective of the training objective, \eg all models associate women more with family roles than careers, and perceive Arab-Muslims as less pleasant than other humans.
This points to the idea that these social biases are indeed ingrained from the training data. 
These findings suggest that biases in image models are not just a result of training data, but that the training objective is a significant factor contributing to their emergence, affecting both the magnitude and direction of biases. 
Hence, we suggest that future work on bias mitigation focuses on the set of social biases that is consistent across models.

\subsection{Impact of Model Architecture}\label{cha:exp-architectural-choices}

\paragraph{Model Size}
The size of a model often impacts its performance, indicating that larger models tend to generate embeddings that contain higher-quality, more general-purpose information about an image.
Therefore, we investigate the influence of model scale on social biases, using iGPT~\cite{iGPT} and ViT-MAE~\cite{ViTMAE}, as both have been trained using self-supervised methods and are available in three different model sizes. 
The results indicate that as we scale the model scales, the direction of social biases within the embedding spaces remains somewhat consistent (see Table~\ref{tab:analysis-model-size}). 
This implies that models of similar architecture, trained on the same dataset using the same training objective, tend to inherit analogous social biases. 

\begin{figure}[!bh]
  \centering
  \small
  \hspace*{-0.8cm}
  \begin{subfigure}{.38\columnwidth}
    \begin{tikzpicture}
\pgfplotsset{%
    width=1.5\textwidth,
    height=1.5\textwidth,
    every tick label/.append style={font=\small}
}
\begin{axis}[
  boxplot/draw direction=y,
  boxplot/variable width=false,
  ymin=0,
  ymax=1.9,
  xticklabels={S, M, L},
  xtick = {1, 2, 3},
  xtick style = {draw=none, font=\small,},
  ytick style={opacity=0, font=\small},
  extra x ticks={2},
  ymajorgrids,
  enlarge y limits,
  y axis line style = {opacity=0},
  axis x line* = bottom,
  axis y line = left,
  extra x tick labels={(a) iGPT},
  extra x tick style={
    yshift=-16pt, %
    xshift=-4pt,
    tickwidth=0 %
  },
  ylabel={Absolute Effect Size $|d|$, see Eq.~\ref{equ:effect-size}},
  y label style={
        yshift=-0.1cm,
        xshift=-0.2cm
    },
]
  \addplot+[boxplot, fill=colorone, draw=linecolor, line width=0.4mm] table[row sep=\\,y index=0] {
    0.777459 \\ 
    0.840396 \\ 
    1.181839 \\ 
    0.905711 \\ 
    0.301433 \\ 
    0.067394 \\ 
    0.455394 \\
    1.219076 \\
    0.975711 \\ 
    0.342669 \\ 
    0.223532 \\ 
    1.305958 \\ 
    1.344787 \\ 
    0.691693 \\
    1.689171 \\
  };
  \addplot+[boxplot, fill=colortwo, draw=linecolor, line width=0.4mm] table[row sep=\\,y index=0] {
    0.381848 \\ 
    0.965109 \\ 
    0.616386 \\ 
    0.458620 \\ 
    0.432178 \\ 
    0.185717 \\ 
    0.074679 \\
    1.019677 \\
    0.468175 \\ 
    0.597871 \\ 
    0.076711 \\ 
    1.257702 \\ 
    0.594381 \\ 
    1.018652 \\
    1.500471 \\
  };
  \addplot+[boxplot, fill=colorthree, draw=linecolor, line width=0.4mm] table[row sep=\\,y index=0] {
    0.395738 \\ 
    0.997152 \\ 
    0.413355 \\ 
    0.788678 \\ 
    0.439345 \\ 
    0.225645 \\ 
    0.265505 \\
    0.606285 \\
    0.769782 \\ 
    0.549317 \\ 
    0.067032 \\ 
    1.109178 \\ 
    0.128592 \\ 
    0.493164 \\
    0.751898 \\
  };
\end{axis}
\end{tikzpicture}
  \end{subfigure}
  \hspace{1cm}
  \begin{subfigure}{.38\columnwidth}
    \begin{tikzpicture}
\pgfplotsset{%
    width=1.5\textwidth,
    height=1.5\textwidth,
    every tick label/.append style={font=\small}
}
\begin{axis}[
  boxplot/draw direction=y,
  boxplot/variable width=false,
  ymin=0,
  ymax=1.9,
  xticklabels={B, L, H},
  xtick = {1, 2, 3},
  xtick style = {draw=none, font=\small,},
  ytick style={opacity=0, font=\small},
  extra x ticks={2},
  ymajorgrids,
  enlarge y limits,
  y axis line style = {opacity=0},
  axis x line* = bottom,
  axis y line = left,
  extra x tick labels={(b) ViT-MAE},
  extra x tick style={
    yshift=-16pt, %
    xshift=-4pt,
    tickwidth=0 %
  },
]
  \addplot+[boxplot, fill=colorfour, draw=linecolor, line width=0.4mm] table[row sep=\\,y index=0] {
    0.10962571948766708\\ 
    0.5524969100952148\\ 
    0.28876230120658875\\ 
    0.34875884652137756\\ 
    0.5877526998519897\\ 
    0.08414085954427719\\ 
    1.1532411575317383\\ 
    1.1506630182266235\\ 
    0.8057575225830078\\ 
    0.3364218771457672\\ 
    0.28753796219825745\\ 
    0.9597402215003967\\ 
    1.3026094436645508\\ 
    1.3105027675628662\\ 
    1.7530592679977417\\
  };
  \addplot+[boxplot, fill=colorfive, draw=linecolor, line width=0.4mm] table[row sep=\\,y index=0] {
    0.03255365788936615\\ 
    0.563248336315155\\ 
    0.20773810148239136\\ 
    0.507177472114563\\ 
    0.5521040558815002\\ 
    0.012056063860654831\\ 
    1.167425513267517\\ 
    1.432068109512329\\ 
    0.7492514848709106\\ 
    0.3544660210609436\\ 
    0.3334307372570038\\
    1.0298738479614258\\
    1.375846266746521\\ 
    1.4116432666778564\\ 
    1.6398261785507202\\
  };
  \addplot+[boxplot, fill=colorsix, draw=linecolor, line width=0.4mm] table[row sep=\\,y index=0] {
    0.09079614281654358\\ 
    0.6297317743301392\\ 
    0.39195936918258667\\
    0.09895025938749313\\ 
    0.5537478923797607\\ 
    0.0937526524066925\\ 
    1.1772112846374512\\ 
    1.3371351957321167\\ 
    0.22624517977237701\\ 
    0.29290100932121277\\ 
    0.3045681416988373\\ 
    0.9545212388038635\\ 
    1.4684538841247559\\ 
    1.4396001100540161\\ 
    0.40454697608947754\\
  };
\end{axis}
\end{tikzpicture}
  \end{subfigure}
  \vspace*{-0.6cm}
  \caption{\textbf{The mean absolute iEAT effect size decreases as model size increases.}  The boxplot illustrates the effect size distribution, with the median (solid line), the quartile range (boxes), and the rest of the distribution (whiskers). \label{fig:boxplots}} 
\end{figure}
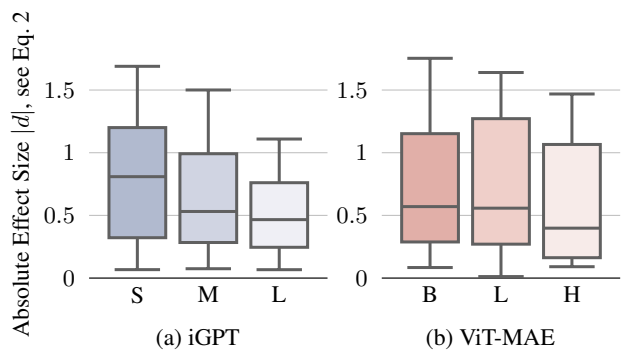

\begin{table*}[!t]
    \small
    \setlength{\tabcolsep}{0.64pt}
    \begin{tabular*}{\textwidth}{l*{15}{x{10mm}}}
        \toprule
            \textsc{Models} & \textsc{T1} & \textsc{T2} & \textsc{T3} & \textsc{T4} & \textsc{T5} & \textsc{T6} & \textsc{T7} & \textsc{T8} & \textsc{T9} & \textsc{T10} & \textsc{T11} & \textsc{T12} & \textsc{T13} & \textsc{T14} & \textsc{T15} \tn
            
        \midrule
        
            \textsc{iGPT-S} & $\textcolor{white}{-}0.66$ & \boldsymbol{$\textcolor{white}{-}0.84$} & \boldsymbol{$-1.02$} & $\textcolor{white}{-}0.75$ & $\textcolor{white}{-}0.22$ & $\textcolor{white}{-}0.16$ & \boldsymbol{$-0.55$} & \boldsymbol{$-1.32$} & $\textcolor{white}{-}0.54$ & $\textcolor{white}{-}0.28$ & $\textcolor{white}{-}0.29$ & \boldsymbol{$\textcolor{white}{-}1.31$} & \boldsymbol{$-1.11$} & $\textcolor{white}{-}0.89$ & \boldsymbol{$\textcolor{white}{-}1.69$} \tn

            \textsc{iGPT-M} & $\textcolor{white}{-}0.38$ & \boldsymbol{$\textcolor{white}{-}0.97$} & $-0.62$ & $\textcolor{white}{-}0.46$ & \boldsymbol{$\textcolor{white}{-}0.43$} & $\textcolor{white}{-}0.19$ & $-0.07$ & \boldsymbol{$-1.02$} & $-0.47$ & $\textcolor{white}{-}0.60$ & $\textcolor{white}{-}0.08$ & \boldsymbol{$\textcolor{white}{-}1.26$} & $\textcolor{white}{-}0.59$ & \boldsymbol{$\textcolor{white}{-}1.02$} & \boldsymbol{$\textcolor{white}{-}1.50$} \tn

            \textsc{iGPT-L} & $-0.40$ & \boldsymbol{$\textcolor{white}{-}1.00$} & $\textcolor{white}{-}0.41$ & $\textcolor{white}{-}0.79$ & \boldsymbol{$\textcolor{white}{-}0.44$} & $\textcolor{white}{-}0.23$ & $\textcolor{white}{-}0.27$ & $-0.61$ & $-0.77$ & $\textcolor{white}{-}0.55$ & $\textcolor{white}{-}0.07$ & \boldsymbol{$\textcolor{white}{-}1.11$} & $\textcolor{white}{-}0.13$ & $\textcolor{white}{-}0.49$ & \boldsymbol{$\textcolor{white}{-}0.75$} \tn

        \midrule

            \textsc{ViT-MAE-B} & $\textcolor{white}{-}0.11$ & \boldsymbol{$\textcolor{white}{-}0.55$} & $-0.29$ & $-0.35$ & \boldsymbol{$\textcolor{white}{-}0.59$} & $\textcolor{white}{-}0.08$ & \boldsymbol{$-1.15$} & \boldsymbol{$-1.15$} & $-0.81$ & $\textcolor{white}{-}0.34$ & $\textcolor{white}{-}0.29$ & \boldsymbol{$\textcolor{white}{-}0.96$} & \boldsymbol{$-1.30$} & \boldsymbol{$-1.31$} & \boldsymbol{$\textcolor{white}{-}1.75$} \tn
            
            \textsc{ViT-MAE-L} & $\textcolor{white}{-}0.03$ & \boldsymbol{$\textcolor{white}{-}0.56$} & $-0.21$ & $-0.51$ & \boldsymbol{$\textcolor{white}{-}0.55$} & $\textcolor{white}{-}0.01$ & \boldsymbol{$-1.17$} & \boldsymbol{$-1.43$} & $-0.75$ & $\textcolor{white}{-}0.35$ & $\textcolor{white}{-}0.33$ & \boldsymbol{$\textcolor{white}{-}1.03$} & \boldsymbol{$-1.38$} & \boldsymbol{$-1.41$} & \boldsymbol{$\textcolor{white}{-}1.64$} \tn

            \textsc{ViT-MAE-H} & $\textcolor{white}{-}0.09$ & \boldsymbol{$\textcolor{white}{-}0.63$} & $-0.39$ & $-0.10$ & \boldsymbol{$\textcolor{white}{-}0.55$} & $-0.09$ & \boldsymbol{$-1.18$} & \boldsymbol{$-1.34$} & $-0.23$ & $\textcolor{white}{-}0.29$ & $\textcolor{white}{-}0.30$ & \boldsymbol{$\textcolor{white}{-}0.95$} & \boldsymbol{$-1.47$} & \boldsymbol{$-1.44$} & $\textcolor{white}{-}0.40$ \tn

        \bottomrule
    \end{tabular*}
    \centering
    \caption{iEAT effect sizes (see Equation~\ref{equ:effect-size}) for a range of association tests (see Table~\ref{tab:ieat-specs}) using different embedding models trained on ImageNet-21k using self-supervised methods. The effect sizes indicate the magnitude and direction of the bias, and are written in bold if the effect is significant at $p_t = 0.05$. \textbf{The direction of the social biases in the embedding spaces of a model are consistent across model sizes. However, the average magnitude of the social biases decreases as model size increases.}\label{tab:analysis-model-size}}
\end{table*}

However, we observe that the average magnitude of the social biases decreases as the model size increases (see Figure \ref{fig:boxplots}), which implies that scaling the model might be a practical strategy to mitigate social biases. 
We speculate that this could be attributed to the model's capacity to capture more semantic information about the objects in the image, without the need to rely on spurious correlations.
However, it is crucial to recognize that scaling a model alone might not be sufficient to eliminate social biases.

\begin{table*}[!b]
    \small
    \setlength{\tabcolsep}{0.75pt}
    \label{tab:resolution-patch}
    \begin{tabular*}{\textwidth}{l*{15}{x{10mm}}}
        \toprule
            \textsc{Models} & \textsc{T1} & \textsc{T2} & \textsc{T3} & \textsc{T4} & \textsc{T5} & \textsc{T6} & \textsc{T7} & \textsc{T8} & \textsc{T9} & \textsc{T10} & \textsc{T11} & \textsc{T12} & \textsc{T13} & \textsc{T14} & \textsc{T15} \tn
        \midrule
            \multicolumn{15}{l}{\textsc{Input Resolution}} \tn

            \textsc{BEiT$_{224}$-L}  & \boldsymbol{$\textcolor{white}{-}1.59$} & \boldsymbol{$\textcolor{white}{-}1.41$} & $\textcolor{white}{-}0.20$ & $-0.07$ & \boldsymbol{$\textcolor{white}{-}0.40$} & $-0.21$ & \boldsymbol{$\textcolor{white}{-}1.59$} & $-0.19$ & \boldsymbol{$\textcolor{white}{-}1.46$} & $\textcolor{white}{-}0.18$ & \boldsymbol{$-0.88$} & \boldsymbol{$\textcolor{white}{-}1.12$} & \boldsymbol{$\textcolor{white}{-}1.06$} & $\textcolor{white}{-}0.81$ & \boldsymbol{$\textcolor{white}{-}1.18$} \tn

            \textsc{BEiT$_{384}$-L} & $\textcolor{white}{-}0.45$ & \boldsymbol{$\textcolor{white}{-}1.46$} & $\textcolor{white}{-}0.60$ & $\textcolor{white}{-}0.15$ & $\textcolor{white}{-}0.36$ & $-0.17$ & \boldsymbol{$\textcolor{white}{-}1.61$} & $-0.46$ & \boldsymbol{$\textcolor{white}{-}1.47$} & $\textcolor{white}{-}0.27$ & \boldsymbol{$-1.11$} & $\textcolor{white}{-}0.47$ & $\textcolor{white}{-}0.61$ & \boldsymbol{$\textcolor{white}{-}1.12$} & \boldsymbol{$\textcolor{white}{-}1.02$} \tn

            \textsc{BEiT$_{512}$-L} & $\textcolor{white}{-}0.01$ & \boldsymbol{$\textcolor{white}{-}1.55$} & $\textcolor{white}{-}0.35$ & $\textcolor{white}{-}0.30$ & $\textcolor{white}{-}0.19$ & $-0.22$ & \boldsymbol{$\textcolor{white}{-}1.65$} & $-0.41$ & \boldsymbol{$\textcolor{white}{-}1.63$} & $\textcolor{white}{-}0.21$ & \boldsymbol{$-1.09$} & $\textcolor{white}{-}0.49$ & $\textcolor{white}{-}0.46$ & \boldsymbol{$\textcolor{white}{-}1.03$} & \boldsymbol{$\textcolor{white}{-}0.79$} \tn

        \midrule
            \multicolumn{15}{l}{\textsc{Patch Size}} \tn
            
            \textsc{DINO-B/8} & $\textcolor{white}{-}0.04$ & \boldsymbol{$\textcolor{white}{-}1.22$} & $\textcolor{white}{-}0.32$ & $\textcolor{white}{-}1.19$ & \boldsymbol{$\textcolor{white}{-}0.37$} & $-0.16$ & $\textcolor{white}{-}0.06$ & \boldsymbol{$\textcolor{white}{-}0.97$} & \boldsymbol{$\textcolor{white}{-}1.16$} & $\textcolor{white}{-}0.36$ & $-0.13$ & $\textcolor{white}{-}0.04$ & \boldsymbol{$-1.21$} & $\textcolor{white}{-}0.41$ & \boldsymbol{$\textcolor{white}{-}1.49$} \tn
            
            \textsc{DINO-B/16} & \boldsymbol{$\textcolor{white}{-}0.99$} & \boldsymbol{$\textcolor{white}{-}1.20$} & $-0.86$ & $\textcolor{white}{-}0.88$ & \boldsymbol{$\textcolor{white}{-}0.38$} & $\textcolor{white}{-}0.01$ & $-0.12$ & \boldsymbol{$\textcolor{white}{-}0.84$} & $\textcolor{white}{-}0.49$ & $\textcolor{white}{-}0.22$ & $-0.08$ & $-0.13$ & $-0.88$ & $-0.77$ & \boldsymbol{$\textcolor{white}{-}1.24$} \tn
        \bottomrule
    \end{tabular*}
    \centering
    \caption{iEAT effect sizes (see Equation~\ref{equ:effect-size}) for a range of association tests (see Table~\ref{tab:ieat-specs}) of BEiT pre-trained on ImageNet-21k and then fine-tuned on ImageNet-1k at different input resolutions, and ViT-DINO trained using different patch sizes. The effect sizes indicate the magnitude and direction of the bias, and are written in bold if the effect is significant at $p_t = 0.05$. \textbf{The direction of the social biases are somewhat consistent across different input resolutions and patch sizes, and the average magnitude of the biases decreases as input resolution increases. However, we do not observe a systematic effect for patch size.}\label{tab:experiment-input-resolution-patch-size}}
\end{table*}

\paragraph{Input Resolution and Patch Size} 
In addition, input resolution and patch sizes have been discussed as important model parameters~\cite{BEiT, ViTMAE}.
Hence, we investigate the effect of these parameters on social biases (see Table \ref{tab:experiment-input-resolution-patch-size}). 
To assess the impact of different input resolutions, we consider BEiT pre-trained on ImageNet-21k at a 224x224 input resolution and subsequently fine-tuned on ImageNet-1k at different input resolutions.
Our results indicate that social biases diminish as input resolution increases. 
This finding implies that adopting higher input resolution might contribute to a reduction in social biases. To assess the impact of different patch sizes, we consider ViT-DINO~\cite{DINO}, which was trained at different patch sizes.
In our analysis, we observe some variability in the magnitude of social biases, but no systematic increase or decrease. 
However, it's important to acknowledge that the sample size for this analysis is small, due to the limited number of published models. 
Therefore, further validation should be conducted to confirm these findings. 

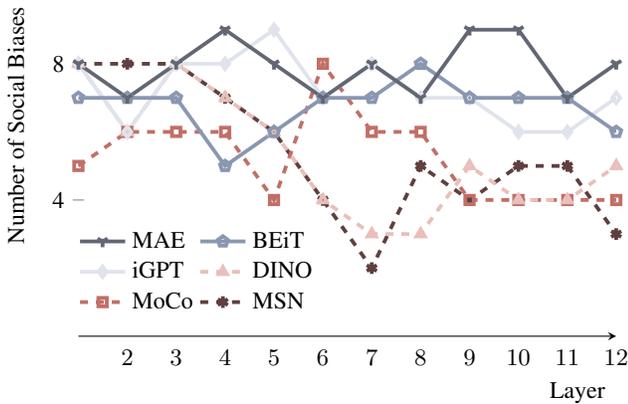
\begin{figure}[!h]
  \centering
  \small%
  \begin{tikzpicture}
\pgfplotsset{
    scaled x ticks=false,
    width=0.5\textwidth,
    height=0.35\textwidth
}
\begin{axis}[
    axis x line=middle,
    axis y line=middle,
    xlabel={Layer},
    ylabel={Number of Social Biases},
    ytick={0, 4, 8},
    yticklabels={0, 4, 8},
    xtick={0, 1, 2, 3, 4, 5, 6, 7, 8, 9, 10, 11, 12},
    xtick={0, 1, 2, 3, 4, 5, 6, 7, 8, 9, 10, 11, 12},
    xtick style = {draw=none, font=\small,},
    ytick style={font=\small},
    legend style={draw=none, fill=none, at={(0.46, 0.34)}},
    legend cell align={left},
    y label style={
        rotate=90,
        yshift=1.05cm,
        xshift=-1.72cm
    },
    x label style={
        yshift=-1cm,
        xshift=-1pt
    },
    legend columns=2,
    ymin=0,
    ymax=10,
    reverse legend,
    y axis line style = {opacity=0},
]

    \addplot[dashed, morecontrast-four, mark=asterisk, mark size=2.2, mark options={solid}, line width=0.5mm] coordinates
     {
     (1, 8)
     (2, 8)
     (3, 8)
     (4, 7)
     (5, 6)
     (6, 4)
     (7, 2)
     (8, 5)
     (9, 4)
     (10, 5)
     (11, 5)
     (12, 3)
     };
     \addlegendentry{MSN} 
    
     \addplot[dashed, morecontrast-five, mark=square, mark size=1.6, mark options={solid}, line width=0.5mm] coordinates
     {
     (1, 5)
     (2, 6)
     (3, 6)
     (4, 6)
     (5, 4)
     (6, 8)
     (7, 6)
     (8, 6)
     (9, 4)
     (10, 4)
     (11, 4)
     (12, 4)
     };
     \addlegendentry{MoCo}

     \addplot[dashed, morecontrast-six, mark=triangle, mark size=1.8, mark options={solid}, line width=0.5mm] coordinates
     {
     (1, 8)
     (2, 6)
     (3, 8)
     (4, 7)
     (5, 6)
     (6, 4)
     (7, 3)
     (8, 3)
     (9, 5)
     (10, 4)
     (11, 4)
     (12, 5)
     };
     \addlegendentry{DINO}

    \addplot[morecontrast-three, mark=diamond, mark size=2, line width=0.5mm] coordinates
     {
     (1, 8)
     (2, 6)
     (3, 8)
     (4, 8)
     (5, 9)
     (6, 7)
     (7, 8)
     (8, 7)
     (9, 7)
     (10, 6)
     (11, 6)
     (12, 7)
     };
     \addlegendentry{iGPT}

     \addplot[morecontrast-two, mark=pentagon, mark size=2, line width=0.5mm] coordinates
     {
     (1, 7)
     (2, 7)
     (3, 7)
     (4, 5)
     (5, 6)
     (6, 7)
     (7, 7)
     (8, 8)
     (9, 7)
     (10, 7)
     (11, 7)
     (12, 6)
     };
     \addlegendentry{BEiT}

     \addplot[morecontrast-one, mark=Mercedes star flipped, mark size=2.2, line width=0.5mm] coordinates
     {
     (1, 8)
     (2, 7)
     (3, 8)
     (4, 9)
     (5, 8)
     (6, 7)
     (7, 8)
     (8, 7)
     (9, 9)
     (10, 9)
     (11, 7)
     (12, 8)
     };
     \addlegendentry{MAE}

\end{axis}
\end{tikzpicture}
  \vspace*{-0.6cm}
  \caption{The number of social biases detected across different embedding layers of ViTs using a significant threshold of $p_t = 0.05$ (see Equation~\ref{equ:test-statistic}). \textbf{ViTs trained using discriminative and generative objectives share a similar number of biases in earlier layers, but diverge as the models form more semantically meaningful features, such that discriminative models encode less social biases in later layers. }\label{fig:per-layer-analysis}} 
  \vspace*{-0.4cm}
\end{figure}

\paragraph{Per-Layer Analysis} 
In our experiments, we use the embeddings from the layer that has been reported to be optimal in linear evaluation. 
However, we expect that the intensity of the biases might differ between layers, due to the increasing semantic interpretability of internal representations~\cite{netdissect2017, StudySSL}.
To explore this, we determine the number of social biases across different layers, using a significance threshold of $p_t = 0.5$.
The results are illustrated in Figure~\ref{fig:per-layer-analysis}.
We observe that for models trained using generative objectives, despite some variation in the magnitude, the number of significant biases is somewhat consistent across all layers.
However, for models trained using discriminative objectives we find that the number of significant biases in the earlier layers mirrors those of models trained using generative objectives and then decreases as we progress through the model. 
This suggests that the biases inherent in the low-level features are consistent across all models, but there is a noticeable divergence as the models develop more semantically meaningful features.
We hypothesize that the observed divergence in biases across different layers could be attributed to the specific training objectives of the models, as detailed in Section 4.2.

The existence of biases in earlier layers does seem counterintuitive, as no semantic concepts have formed yet. 
However, we found a substantial portion of these biases, such as skin tone and weight, are connected to lower-level features, such as pixel brightness. 
This suggests that these biases could be identified without necessarily associating them with the intended semantic concepts.
Therefore, we hypothesize that the root of the biases in the earlier layers could be grounded in the inherent characteristics of the image data, and not necessarily the high-level semantic interpretations we are probing.
These findings align with prior observations on ResNets~\cite{StudySSL}.

\section{Conclusion}
The emergence of social biases in models trained using self-supervised objectives is often attributed to biases in the training data. 
However, we find that models can exhibit opposite biases despite being trained on the same data.
This challenges the prevailing belief that social biases arise just from simple co-occurrences of objects in the training images. 
Moreover, we find that training objectives, model architecture, and model scale each have significant effects on social biases in learned representations. 
These effects can be the reduced, but not eliminated, using counterfactual data augmentation. 
Therefore, we recommend that model developers and users take these details into account in designing and selecting the model most relevant to their needs, as each decision has quantifiable trade-offs.
Moreover, our analysis exposes a set of social biases that is consistent across different models, wherefore we suggest that future work assesses their bias mitigation approaches on these dimensions. 

\section*{Acknowledgment} 
This work was supported in part by the German Federal Ministry for Digital and Transport (BMDV), and in part by the German Federal Ministry for Economic Affairs and Climate Action (BMWK).

{\small
\bibliographystyle{ieee_fullname}
\bibliography{egbib}
}
\end{document}